**Using Known Words to Learn More Words:**

**A Distributional Model of Child Vocabulary Acquisition**

Andrew Z. Flores, Jessica L. Montag, & Jon A. Willits

Department of Psychology

University of Illinois at Urbana Champaign

**Abstract**

Why do children learn some words before others? A large body of behavioral research has identified properties of the language environment that facilitate word learning, emphasizing the importance of particularly informative language contexts. However, these findings have not informed research that uses distributional properties of words to predict vocabulary composition. In the current work, we introduce a predictor of word learning that emphasizes the role of prior knowledge. We investigate item-based variability in vocabulary development using lexical properties of distributional statistics derived from a large corpus of child-directed speech. Unlike previous analyses, we predicted word trajectories cross-sectionally, shedding light on trends in vocabulary development that may not have been evident at a single time point. We also show that regardless of a word's grammatical class the best distributional predictor of whether a child knows a word is the number of other known words with which that word tends to co-occur.

## 1. Introduction

One of the fundamental challenges of early language learning is the ambiguity involved in learning word-referent relationships. For instance (in a variant of Quine's classic 1960 example), if a child hears someone say "bunny" in a pet shop, the child may not know which of the unfamiliar animals to which the word refers, or even that it refers to an animal at all. The person might have been saying "cute!" or "look!" or "smelly!". How do language learners overcome the problem of a vast search space, and learn the correct mapping between words and referents?

Learning new words is a complex process, and many word learning studies have examined how basic learning mechanisms and inductive biases promote vocabulary growth. Learning mechanisms such as fast mapping (Carey & Barlett, 1978), analogical learning (Gentner, 1989), cross-situational statistical learning (Yu & Smith, 2007), distributional learning (Gleitman, 1990; Harris, 1957; Lany & Saffran, 2010), and hypothesis testing (Trueswell, Medina, Hafri & Gleitman, 2013) all provide means by which a learner can develop reasonable hypotheses about word-referent mappings. In turn, inductive biases may simplify the problem by reducing the number of hypotheses a learning mechanism needs to actively consider. Inductive biases such as the mutual exclusivity principle (Markman & Wachtel, 1998) suggest language learners tend to associate only a single label per object, which prevents word-referent mappings from overlapping. Likewise, a bias to assume that objects will have the same label if they are similar in shape (but not if they are similar in color or texture) suggests that language patterns tend to associate labels with items of the same shape, again simplifying the label-to-object mapping problem (Smith et al., 2002). Other biases have also been proposed, such as those that focus on children's tendency to pay attention to particular social cues like eye-gaze and pointing (Akhtar, Carpenter & Tomasello, 1996; Tomasello, 1983, 1988; Yu & Ballard, 2007). Understanding the learning mechanisms and inductive biases that allow children to learn language has been a major goal of the field of language development.

How children learn to map words to references has typically been studied two ways. First, experimental studies of word learning in controlled laboratory settings have tested many hypotheses about learning mechanisms and inductive biases that may scaffold the learning process. Second, statistical approaches, facilitated by the availability of large development datasets, have allowed researchers to investigate which factors predict language outcomes (Goodman, Dale & Li, 2008; Frank, Braginsky, Yurovsky, Marchman, 2021). In these kinds of studies, large datasets of surveys of parents that assess whether their children understand or produce specific words are used as outcome variables, and attempts are made to find correlational predictors of these outcome variables. Both experimental and statistical approaches have contributed to our understanding of how the language learning process unfolds, with each method providing new information, as well as raising new questions.

Reflecting on this work, it is clear that many insights from the behavioral work have not yet been incorporated into the statistical modeling approach. In the current work, we highlight three insights from behavioral word learning experiments that we believe can inform statistical approaches. Incorporating these findings from behavioral work may increase the predictive power and ecological validity of the statistical models that aim to describe early word learning. The first of these insights is the relationship between grammatical class and vocabulary development, and the extent to which there are class-specific biases and learning mechanisms. The second insight is the distinction between quantity and quality in linguistic experience, and specifically the question of what constitutes a "high quality" learning episode. The third insight is the incorporation of information about prior knowledge children have about other words into prediction of subsequent word learning, and the way in which language acquisition is an interactive process with many top-down effects.

In order to address the absence of these factors from statistical models of vocabulary development, we introduce a new predictive measure of word learning that incorporates insights from behavioral research, focusing on these three topics. The innovation of our approach is that this predictor uses children's prior knowledge as a means of specifying one way in which a learning episode can constitute a "high quality" learning episode. And unlike other statistical approaches to word learning, this prior-knowledge-based predictor eliminates the need to posit that a word's grammatical category affects the mechanisms by which that word was learned.

**The role of grammatical class in word learning.**

**Behavioral Evidence.** Most of the previously discussed learning mechanisms (such as fast mapping, cross-situational learning, and hypothesis testing), as well as inductive biases (such as mutual exclusivity, shape bias, and social cues) have generally focused on the learning of nouns. However, most of these learning mechanisms and inductive biases have also been proposed to be applicable in a more general way, to words from multiple grammatical classes. Continued behavioral research is actively exploring the extent to which these learning mechanisms extend to other parts of speech, or whether lexical-class-specific learning mechanisms and inductive biases are necessary. A number of these mechanisms are explicitly proposed to be general mechanisms, applying to words in any grammatical class. For example, statistical and distributional learning mechanisms seem to be applicable to multiple grammatical classes. The distributional statistics of a word's prosodic information, word co-occurrence information, and syntactic information have each been shown to be useful for inferring aspects of meaning of words from multiple grammatical classes (Arias-Trejo & Alva, 2013; Christophe et al., 2008; Fisher, Gertner, Scott, Yuan, 2010; Hills, Maouene, Riordan & Smith 2010; Lany & Saffran, 2010, 2013; Naigles, 1990; Wojcik and Saffran, 2013). Indeed, formal computational models of distributional semantics tend to not make *a priori* distinctions between grammatical classes, and perform well at learning thematic and taxonomic relations across many grammatical categories (Burgess & Lund, 1996; Elman, 1990; Huebner & Willits, 2018; Jones & Mewhort, 2007). Likewise, most proposals involving analogical learning, Bayesian inference, and hypothesis testing that are formally applicable to the word learning process have been shown to apply to learning about the aspects of meaning of words from multiple grammatical classes (Booth & Waxman, 2009; Gentner, 1989; Gentner & Namy, 2006; Sadeghi, Scheutz, Krause, 2017).

In short, while most learning mechanisms and inductive biases have been demonstrated in the context of noun learning (Akhtar et al., 1996; Carey, 1978; Markman, 1998; Smith et al., 2002; Tomasello, 1983, 1988; Trueswell et al., 2013; Yu &

Smith, 2007), they are hypothesized to be at least partially independent of the grammatical class of the word that is being learned, despite nouns being the demonstrated test case.

**Statistical Models.** Statistical models of word learning typically investigate multiple lexical categories, and are used to look for predictors of differences in language learning outcomes across individuals or across words. Some of these studies have focused on children as the random variable, looking for predictors of which children are likely to have high vocabularies, and which children are likely to have low vocabularies. These studies - which include both correlational studies and statistical models or regression models - have found that many demographic factors, such as gender (Huttenlocher et al., 1991), maternal education (Pan, Rowe, Singer & Snow, 2005), birth order (Havron et al., 2019), amount of language input (Hoff, 2003; Huttenlocher et al., 1991; 2010; Weisleder & Fernald, 2014), and lexical processing speed (Hurtado, Marchman & Fernald, 2008), are all predictors of higher vocabularies. Other studies have focused on the words as the random variable, looking for predictors of which words are likely to be understood or spoken at earlier ages, and which words are likely to be understood or spoken at later ages. These studies have found that many distributional and semantic properties of words are predictive of an earlier mean age of acquisition, including word frequency (Blackwell, 2005; Braginsky et al., 2021; Goodman et al., 2008; Naigles & Hoff-Ginsberg, 1998), contextual diversity (Blackwell, 2005; Hills et al., 2010; Hsu, Hadley & Rispoli, 2017; Naigles & Hoff-Ginsberg, 1998), concreteness (Braginsky et al., 2021; Swingley & Humphreys, 2018), positive valence (Braginsky, Yurovsky, Marchman and Frank, 2019; Moors et al., 2013), and child "relevance" (Perry, Perlman, Winter, Massaro & Lupyan, 2018).

In contrast with the learning mechanisms proposed and tested in behavioral experiments, the distributional and semantic predictors of a word's age of acquisition are very much *not* independent of grammatical class. Many investigations focus on a single grammatical class of word (e.g., adjectives: Blackwell, 2005; verbs: Hsu, Hadley & Rispoli, 2017; Naigles & Hoff-Ginsberg, 1998). Further, a word's grammatical class is itself a strong predictor of its age of acquisition, with nouns acquired before verbs, and verbs acquired before adjectives, and adjectives acquired before function words (Fenson, et al., 1994; Swingley & Humphrey, 2018; Gentner, 1982).

Even more striking is that all of the distributional predictors are themselves dependent on grammatical class when multiple classes are investigated at the same time. For example, the correlation between word frequency and a word's age of acquisition has been shown to depend on grammatical class (Braginsky et al., 2021; Goodman et al., 2008). Specifically the relationship between frequency and age of acquisition is non-significant (or even negative) when examined across all words, but is significantly positively correlated within words of a specific grammatical class. The strength of the effect of frequency within each class varies as well, with the effect of word frequency being quite strong for nouns, and smaller (though still significant) for verbs, adjectives, and function words (Goodman et al., 2008). Likewise, predictors such as contextual diversity which measure the count of unique words or contexts with which a given word co-occurs, shows a similar sensitivity to word class. Hills et al. (2010) showed that when controlling for the effect of a word's frequency, words that co-occur with a greater number of words in the language environment predicts subsequent word learning, but with considerable variation for different grammatical classes. Findings like these suggest that different grammatical categories may be learned via different learning mechanisms (Hills et al, 2010). Across many studies, statistical models of word learning imply, either implicitly (by only investigating a single grammatical class) or explicitly (by finding different statistical predictors of word learning across grammatical classes) that statistical predictors of word learning are best interpreted in conjunction with information about grammatical class.

A final example of distributional statistical approaches not being independent of grammatical class is work by Chang and Deak (2020). Chang and Deak created a two word co-occurrence matrices (one with content words within sentences, and one with immediately adjacent syntactic frames), and then computed the principle components of those matrices. They then used words' loadings on these principle components as predictors of the age at which a threshold number of children comprehend or produce a word. Chang and Deak found that many of these principle component loadings predict MCDI extremely well. Notably many of the most highly predictive principle components where grammatical in nature. For example, the strongest effect came from principle component 2 of the syntactic frame co-occurrence matrix, which more or less indexed whether word was a noun versus a function word.

The fact that statistical models of vocabulary development based on distributional information require (or at least benefit from) information about grammatical class, stands in stark contrast to the proposals put forth in behavioral studies, which suggest class-independent learning mechanisms. This mismatch suggests one of two conclusions. One possibility is that implications of the statistical work are being undervalued, and the grammatical class-agnostic learning mechanisms like distributional learning are not as independent of grammatical class as they seem. Alternatively, the mismatch could be due, not to the fact that no grammatical class independent distributional predictor exists, but that none has yet been found. In either case, resolving this inconsistency between the behavioral and statistical research would shed considerable light on mechanisms of vocabulary acquisition.

**The quality and quantity distinction**

**Behavioral Evidence.**

Within the language acquisition literature, there is often a distinction made between the quantity and quality of speech that children hear, with different proposals about the role that both quantity and quality of experiences play in language development. There is substantial evidence that both language quantity and quality are associated with language outcomes. Higher quantities of speech to children are associated with positive language outcomes (Dickinson & Tabors, 1991; Shneidman, Arroyo, Levine, Shneidman et al., 2013; Hart & Risley, 1995; Hoff, 2003; Huttenlocher et al., 2010; Song et al., 2012; Weisleder & Fernald, 2014). While effects of input quantity alone may well predict outcomes, many researchers believe that what matters most is not the number of times a child hears a word, but rather the number (or proportion) of times they hear that word in a "high quality" context (Tomasello, 1998; Hirsh-Pasek et al., 2015; Hoff, 2006; Yu & Smith, 2007).

So while greater quantity may provide more opportunities for high quality language learning episodes, it is the linguistic quality itself, and what makes a context high or low quality, that has been the focus of research. For example, social contexts that reduce referential ambiguity are thought to contribute to high quality contexts. One factor shown to be predictive of a high quality learning episode is reduction of referential ambiguity by the use of socio-visual cues (Cartmill, Armstrong, Gleitman, Goldin-Meadow, Medina & Trueswell, 2013). Another is whether caregiver and child engage in joint attention (Tomasello, 1983, 1988; Akhtar et al., 1996) or whether the child's attention is sustained on the target item (Yu, Suanda & Smith, 2019).

In addition to identifying instances where referential ambiguity is reduced, a great deal of research has focused on discovering properties of the language input itself that contribute to high-quality learning episodes. Caregiver speech has been shown to possess various lexical and syntactic qualities that aid language development, such as lexical diversity (Hoff & Naigles, 2002; Huttenlocher et al., 2010; Pan et al., 2005; Rowe, 2012), syntactic complexity (Cameron-Faulkner, Lieven, & Tomasello, 2003; Huttenlocher et al., 2002; Rowe, Leech & Cabrera, 2017), or speech that is particularly sensitive or responsive to the child's behavior (Harris, Jones & Grant, 1983; Hirsh-Pasek et al., 2015; Tamis-Lemonda, Kuchiro & Song, 2014). Likewise, speech that is child-directed rather than overheard by the child is particularly associated with positive outcomes (Shneidman et al., 2013; Weisleder & Fernald, 2014), as are word contexts that are particularly informative of word meanings, or relationships between multiple concepts (Beals, 1997; Rowe, 2012). Finally, variability in the contexts in which words appear aids generalization of word labels to new exemplars (Goldenberg & Sandhofer, 2013; Vlach & Sandhofer, 2011).

Certain high-quality contexts also facilitate children's ability to segment words from fluent speech. Reliable cues or anchors that occur in highly familiar or frequent contexts help young language learners reduce potential candidates for new words in fluent speech. These include high frequency lexical items like a child's name (Bortfeld, Morgan, Golinkoff & Rathbun, 2005), highly frequent functional morphemes that reliably precede nouns (Shi & Lepage, 2008) and frequent contextual frames that tend to co-occur with nouns and verbs (Willits, Seidenberg & Saffran, 2014).

To summarize, considerable behavioral research has been focused both on the quantity of input that a child receives, and also on many aspects of the quality of those experiences. Though effects of frequency clearly appear across multiple dimensions of language learning (Ambridge, Kidd, Rowland & Theakston, 2015), a theme that emerges in the behavioral literature is that it is the quality of experiences, and not necessarily the raw quantity alone, that is more important for predicting language outcomes (Anderson et al., 2021; Hirsh-Pasek et al., 2015).

**Statistical Models.** Statistical models have also attempted to address the question of the relative importance of quantity versus quality in predicting vocabulary development outcomes. It might seem like this is a relatively straightforward question to investigate, as word frequency (the number of times a child hears a word) can be considered a proxy for quantity. This can then be contrasted with other distributional predictors that are used to operationalize various hypotheses of the qualitative aspects of learning episodes. For example, researchers have investigated whether earlier learned words are special in terms of their lexical contextual diversity (the number of other words with which a word co-occurs, Blackwell, 2005; Hills et al., 2010; Naigles & Hoff-Ginsberg, 1998), and episode diversity (the number of different episodes in which a word occurs, Harris, Barrett, Jones & Brookes, 1988; Roy, Frank, DeCamp, Miller & Roy, 2015). These studies have tended to find that higher lexical diversity and lower episodic diversity are associated with earlier learned words. Similarly, researchers have found that words that occur more frequently in isolation tend to be learned earlier (Brent & Suskind, 2001), as do words that more frequently occur in shorter utterances (Swingley & Humphrey, 2018) and words that occur more frequently at the beginning and ending of utterances (Braginsky, Yurovsky, Marchman & Frank, 2016).

From one perspective, the statistical literature seems to parallel the behavioral literature quite closely, with both quantity and many measures of quality each predicting language learning outcomes. But in fact, predictors that are more associated with "quality" often have very small effect sizes, or even go away, when simple word frequency is controlled. For example, in Braginsky et al.'s (2019) study investigating many distributional statistics' ability to predict MCDI scores in English, word frequency's effect size was approximately $r = 0.45$, while the effects of appearing in short utterances, appearing alone, and appearing at the end of an utterance were approximately $r = 0.30$, $r = 0.15$ and $r = 0.03$, respectively.

As in the case of grammatical class, the relative contribution of linguistic quantity versus quality shows a discrepancy between the behavioral and statistical research, with the behavioral research emphasizing quality learning episodes, while the statistical research routinely shows that quantity (word frequency) to be the best predictor of easy-to-learn and hard-to-learn words (c.f., Roy et al., 2015). This mismatch between the behavioral and statistical literatures again suggests one of two conclusions. One possibility is that once again the implications of the statistical work are being undervalued, and along with it the importance of pure quantity as an important factor in vocabulary acquisition. Alternatively, the mismatch could again be pointing to the failure of the statistical work to correctly identify, measure, and use adequate proxies for high quality learning episodes. As before, resolving this inconsistency between the behavioral and statistical research would shed considerable light on mechanisms of vocabulary acquisition.

**Prior Knowledge**

**Behavioral Evidence.**

One extremely important contribution of experimental word learning research has been the demonstration of a wide range of ways that word learners use preexisting knowledge of other words to bootstrap the learning of new words. Each word learning episode is not independent or done in isolation, and both general learning mechanisms and inductive biases take advantage of prior knowledge. The role that prior knowledge plays in driving subsequent learning is a central theme in the word learning literature.

There are many examples of this phenomenon outside of the lexical level. For example, infants' sensitivity to the distributional structure of the sounds in their language affects their phonemic discrimination (Maye, Werker, & Gerken, 2002), and infants have an easier time recognizing, processing, and learning new syntactic structures that match those with which they have previous experience. For example, nonadjacent dependency learning is bootstrapped by prior learning of an adjacent dependency (Lany, Gomez, & Gerken 2007), when the dependencies share phonological overlap (Onnis, Monaghan, Richmond & Charter, 2005), semantic overlap (Willits, Lany, & Saffran, 2020), or are cued by known nonadjacent dependencies (Zettersten, Potter, & Saffran, 2020). Likewise, children's ability to produce and understand complex sentences, like those that contain relative clauses, seems to emerge from children's ability to use and understand simpler sentences (Brandt, Diessel & Tomasello, 2008), and are also predicted by whether the child has a high or low vocabulary (Borovsky, Elman, & Fernald, 2012; Fernald, Marchman, & Weisleder, 2013).

Within the realm of learning about words and their meanings, there is a tremendous amount of evidence that children and adults lean heavily on pre-existing knowledge while learning new words. Specifically, by using prior knowledge, children are able to narrow down potential referential candidates of new words. For example, children are able to use known object labels to reduce referential candidates through the principle of mutual exclusivity (Merriman, Bowman & MacWhinney,1989; Markman, 1998), and through comparing prior experiences to new ones discovering common abstractions though analogical learning (Gentner, 1989).

Further, there is a great deal of evidence that children keep track of and accumulate knowledge of statistical regularities in the language environment, and can use this information to aid word learning. Studies which examine children's capacity to learn from statistical regularities have shown they can make inferences about a word's semantic category as a result of their patterns of distributional co-occurrence (Lany & Saffran, 2010). Children can also use known nouns to infer meanings of verbs (Landau & Gleitman, 1985; Naigles, 1996; Yuan & Fisher, 2009) and known verbs to infer meanings of nouns (Ferguson, Graf & Waxman, 2014). Research with ERPs shows that new words are learned more easily when they occur in semantically supportive contexts (Borovsky, Kutas, & Elman, 2010), and patterns of word co-occurrence and category structure differentially predict differences in word learning for high and low vocabulary children (Borovsky & Peters, 2019). Children are also able to apply previous encounters with distributional regularities, such as when children are tasked with rapidly evaluating statistical evidence across individually ambiguous words to resolve word-referent ambiguities in cross-situational learning tasks (Yu & Smith, 2007). Children are also able to display prior knowledge of the sounds and phonotactics of their language to aid in other tasks. Children more easily recognize novel words that follow their native language's phonotactic (Jusczyk & Aslin, 1995; Nazzi et al., 2005) and stress (Echols, Crowhurst, & Childers, 1997; Houston, Santelmann, & Jusczyk, 2004; Jusczyk, Houston, & Newsome, 1999; Morgan & Saffran, 1995; Nazzi et al., 2005) patterns. Prior experience with phonological forms also assists individuals with mapping novel word forms to references (Estes, Evans, Alibali, & Saffran, 2007; Fennell & Werker, 2003; Ferry, Hespos & Waxman, 2010; Hay, Pelucchi, Graf Estes, & Saffran, 2011). Similarly, infants can recognize and attend to the visual referent of a word at much earlier ages if it is spoken by a familiar voice, such as their mother (Bergelson & Swingley, 2012).

It is then evident that many forms of prior knowledge that children bring to word learning tasks, including acoustic, lexical and syntactic knowledge, aids in the language learning process. Despite a rich literature citing the importance of prior knowledge for subsequent learning, prior knowledge has often not been incorporated into statistical models of word learning, or has been incorporated in narrow ways.

**Statistical Models.** Despite the widespread acceptance and considerable work showing that prior knowledge is important for understanding word learning in behavioral studies, prior knowledge is rarely incorporated into statistical models of word learning.

The notable exception to this is work by Hills and colleagues, who used a growth model analysis in graphical networks to simulate how new words might be added to the network. In Hills et al. (2010), 15 separate graphs were produced for each month representing children aged 16 to 30 months, with nodes added to the graph for each word produced by at least 50% of children at that age (according to MCDI parental surveys). Links between nodes were added if the words ever co-occurred within a sentence in the CHILDES database (using a version of the corpus containing approximately 2 million words of speech from children aged 12 to 60 months). They then calculated the connection density of each word at each age (the *indegree*), by calculating how many connections the word had. They used this measure of degree of connection density to predict which words were added at each age. They tested three hypotheses about how connectivity predicted the acquisition of new words. The "preferential attachment" hypothesis was that the new words that were most likely to be added were words that co-occurred with already known words (i.e. words already in the network) that had the highest degree, or the densest connectivity with other known words. The "preferential acquisition" hypothesis was that words that co-occurred with the most unknown words would be the first to be added (in other words, the most densely connected of the words that hadn't yet been learned). The "lure of the associates" hypothesis is that an unknown word that occurred with the most words already in the network would be the first to be learned. Hills et al. found that the "lure of the associates" hypothesis best overall word acquisition, and also best predicted noun acquisition. But they found that verbs and function words were best predicted by the "preferential acquisition" hypothesis, and that none of the hypotheses predicted the acquisition of adjectives.

Hills et al.'s analyses are interesting and notable because they are one of the few studies that attempted to take prior knowledge into account when predicting vocabulary development. But the study also raises many questions. Taken at face value, these findings resurface the issue of the independence of grammatical class from word learning. Are qualitatively different learning mechanisms at work for different grammatical classes? There is also a question of how to compare the results from Hills's analyses to the kinds of large-scale regression analyses that predict individual child level data (as opposed to the graphical growth models where nodes are added when more than 50% of children say a word). A final question is the validity and interpretability of using binary co-occurrence values in a corpus for children up to age five to predict the month-by-month acquisition of new words, and whether it makes sense to use these binary co-occurrences counts as measures of the associativity of words already in the network, outside the network, and between words already in and not yet in the network.

But aside from the exception of Hills et al., statistical models of word learning compute measures of distributional statistics directly from corpora alone, so there is little opportunity for features of words that do not directly arise from the corpus itself - such as the likelihood that children know that word - to be incorporated into statistical models of word learning. We believe that incorporating measures of prior knowledge into distributional measures of the language environment may yield better estimates and predictors of word learning.

**The Present Study**

From a brief review of the behavioral literature and the statistical modeling literature, we have noted three features that emerge that typically distinguish the processes thought to underlie word learning in the behavioral and modeling literatures. First, behavioral theories are typically robust to grammatical class, whereas statistical models often find different effects of statistical predictors for different grammatical classes. Second, in the behavioral literature, measures of linguistic quality, broadly defined, are typically investigated as key predictors of language outcomes, and typically outperform simple quantity of linguistic input on various metrics. In statistical models, measures of linguistic quantity, specifically, word frequency, are typically the best predictors of word learning. Third, behavioral investigations of word learning have extensively documented the ways in which new word learning builds on a child's prior language knowledge, but statistical models of word learning rarely incorporate prior knowledge into subsequent word learning. We believe that by incorporating these three features that emerge from behavioral studies into statistical models, we can develop a statistical predictor of word learning that both better predicts children's word knowledge than existing statistical measures and incorporates key findings from behavioral research into statistical models.

We introduce a predictor variable that is designed to capture how children's existing word knowledge interacts with information present in the distributional properties of words, the Proportion of Known Words with which a word co-occurs (or Pro-KWo). Pro-KWo (our operational definition of which is described in greater detail in the Methods section) differentiates words that tend to co-occur with more words that children are likely to know, versus words that co-occur with fewer words that children are already likely to know. If prior word knowledge indeed contributes to subsequent word learning, words assigned a high score on this measure should in principle be easier to learn and produce. The effect of prior knowledge, specifically, the co-occurrence of a word with known words, may yield particularly useful word learning contexts and be an aspect of linguistic quality that can be incorporated into statistical models of word learning. To give an intuitive example of Pro-KWo, consider the words "where" and "why." One contributing factor to why children may produce "*where*" before "*why*" is that "*where*" tends to co-occur with words children already know and whose location is getting asked about. In contrast, "why" is often part of questions that involve less frequent, and more abstract, and therefore later-learned referents. A distributional learner should take longer to acquire the word "*why*" because the meanings of the words that co-occur with "*why*" are themselves less likely to be known.

Pro-KWo is thus a cousin of measures like lexical contextual diversity. A word has high lexical contextual diversity if it occurs with a higher proportion of total word types. But lexical contextual diversity does not consider whether those co-occurrences are with known or unknown words, whereas a word will have a high Pro-KWo score if the total proportion of its co-occurrences is likely to be with words a child already knows.

Thus, Pro-KWo is conceptually more similar to the "lure of the associates" measure proposed by Hills et al. (2010), though operationally there are several important differences in how we have defined our Pro-KWo measure that makes it different from Hills et al.'s "lure of the associates" hypothesis. A word's "lure of the associates" score depends on its structural connectivity within a graphical lexical network; a word scores high if it is connected to more words that are already known. In our measure, it is the proportion of overall co-occurrences that matters. Thus, in "lure of the associates", a word that co-occurs with five known words is more likely to be acquired than a word that co-occurs with three known words; for Pro-KWo, the latter word could be predicted to be the earlier-acquired word if the word's ratio of co-occurrences with those three words (relative to its co-occurrences with all other words) is extremely high. Thus words with a high Pro-KWo score may or may not co-occur with many different words, they may not even co-occur with many different known words; the important thing is that a high proportion of a word's occurrences are in already known contexts. Thus, Pro-KWo is designed to be a distributional analogue of the behavioral research demonstrating that prior knowledge often aids word learning through mechanisms that rely on children to already know some of the other words in the sentence (whether those mechanisms be bootstrapping mechanisms like syntactic bootstrapping, or constraint-based mechanisms like mutual exclusivity).

Building upon Hills et al (2010) we explore a novel way in which prior knowledge can be incorporated into statistical models of word learning. We believe this method may be a way to both include key findings from behavioral experiments into statistical models of word learning, and improve the accuracy of statistical models that predict word learning. The open question is whether the Pro-KWo measure, like "lure of the associates", also shows strong interactions with grammatical class, or is independent of it.

## Method

In order to predict the words that children know from distributional statistics of child-directed speech, we must first operationalize and compute both measures of child vocabulary and the four key distributional statistics describing patterns in child-available speech, including our new Pro-KWo measure.

### Dependent Measures: Child Vocabulary Data and MCDlp.

First, we operationalize word knowledge as children's word production tracked as part of the American English MacArthur-Bates Communicative Development Inventory (Fenson, 2007). To predict word production data, we use multiple predictors computed from distributional statistics from child-available speech in the American English CHILDES corpus (MacWhinney, 2000), including our new measure, Pro-KWo.

The words used in our analyses are the 680 items from the American English MacArthur-Bates Communicative Inventory of child language production (Fenson, 2007). We obtained the results of MCDI (Words & Sentences) surveys for 5,520 parents, available at the Wordbank website (http://wordbank.stanford.edu, Frank et al., 2017) which reports whether or not a child produces a word at a given age. The data was downloaded on May 22, 2021 using the R *wordbank* package (http://wordbank.stanford.edu/). In our analysis, we excluded duplicate homonyms (i.e "can"), compound words (i.e "french fries") , word endings (e.g "eat-ing") and specific items (a list of all excluded items can be found at: *https://github.com/AzFlores/Pro-KWo*). The final dataset included 499 words.

For our outcome variables, we used two measures of children's vocabulary knowledge. The first, hereafter MCDIp (MCDI proportion), refers to the proportion of children who produced a particular word at each age. To calculate a word's MCDIp score, we first summed the number of times a word is reported as produced in the MCDI, then we divided that sum by the total number of administrations. This procedure yielded 499 individual MCDIp scores (one for each word) for each age from 16 to 30 months. Therefore, values for each word increase with age (but not always) as more children produce each word. Similar analysis of children's vocabulary development has used MCDI values to estimate a word's age of acquisition by calculating when a given word exceeds a threshold (Braginsky et al., 2017; Goodman et al., 2008; Hills et al., 2010).

Our second dependent measure includes the binary production outcomes (child produced or did not produce a word) for all MCDI surveys from age 16-30 months.

**Predictors using Distributional Statistics**

All lexical distributional statistics used as predictor variables in our analyses were derived from the CHILDES database, a corpus of speech addressed to and in the presence of children (MacWhinney, 2000). Our dataset includes 49 corpora of American English spoken to 522 children up to 30 months of age. The data was obtained from the Childes-db website ([http://childes-db.stanford.edu](http://childes-db.stanford.edu), Sanchez et al., 2018) on May 22, 2021, using the R package *childesr* (Braginsky, Sanchez & Yurovsky, 2018). Using this dataset, we obtained the distributional statistics for the 499 MCDI words as described below.

**Cumulative Log Frequency (Frequency).** Each MCDI word's log frequency was computed by counting the number of times it occurred in the CHILDES corpus for children up to a given age, and then performing a $\log_{10}$ transformation. This resulted in 15 log frequency scores for each word, one for each age between 16-30 months.

**Lexical Diversity (LD).** Lexical diversity was computed by counting the proportion of other MCDI words that each MCDI word co-occurs with. This was computed in a manner similar to the Hyperspace Analogue to Language (HAL) model (Lund & Burgess, 1996). First, for each age from 16-30 months, we constructed a 499x499 matrix, with each cell in the matrix reflecting the number of times each word co-occurred with another MCDI word in the CHILDES corpus within a 7-word (forward) window. This resulted in 15 (one for each age) different 499-element co-occurrence vectors for each MCDI word. For each age, we then computed the proportion of each word's vector elements that were nonzero, to obtain the proportion of MCDI word types that each word co-occurred with at that age.

**Document Diversity (DD).** Document diversity was calculated by computing the proportion of the 1718 documents (i.e number of transcripts in our CHILDES dataset) in which a word occurred. Each individual audio recording (document) in CHILDES captures a single event such as breakfast or bathtime, so document diversity can be considered a proxy of the diversity of events in which a word occurs. This resulted in 15 document diversity scores for each MCDI word (one for each age group), the number of documents for each word is cumulative.

**Proportion Known Word Co-occurrence (Pro-KWo)**. Our measure of the "Proportion of Known Word Co-occurrence" (Pro-KWo) , was computed as follows. We started with the HAL-like matrix described above when computing contextual diversity. This matrix yielded counts of how many times each MCDI word co-occurred with each other MCDI word. We then took each word's 499-element co-occurrence vector, and multiplied those values element-by-element by the MCDIp score for each co-occurring word, which served as a proxy for how likely it is that children of that age produced each word. This yielded, for each word at each age, a 499-element vector of co-occurrence frequencies, weighted by the proportion of children who know each of those 499 co-occurring words. Next, we calculated each vector's sum of both the original unweighted (HAL-like) counts and the counties weighted by the MCDIp. We divided the *unweighted* sum by the corresponding *weighted* co-occurrence vector. The resulting scalar value is a proxy for the proportion of a word's total co-occurrences with produced words. An example is shown in Table 1 using hypothetical but illustrative MCDIp scores and co-occurrence counts. This table shows that the words why and where, while equated in frequency, nonetheless have very different Pro-KWo scores because "where" co-occurs with more known words.

**Pro-KWo Shuffle.** Our measure of Pro-KWo is based on an aggregate measure of children's vocabulary knowledge, because each MCDIp value is an average proportion of children who produce a word below some age. Given the propensity of MCDIp scores to be more similar within age groups, we wanted to exclude the possibility that age was confounded with Pro-KWo. To control for age, we created a Pro-KWo Shuffle measure. First we shuffled mcdi scores 1000 times and used each instance to calculate 1000 individual Pro-KWo scores (for each word, at each age). We then correlated each newly created shuffled Pro-KWo score with the non-shuffled MCDI and computed the average correlation at each age.

| 1. Unweighted co-occurrence counts. | | | | | |
|---|---|---|---|---|---|
| | ball | cup | think | did | Sum |
| **Why** | 10 | 10 | 100 | 100 | 220 |
| **Where** | 100 | 100 | 10 | 10 | 220 |
| *MCDIp* | *0.7* | *0.6* | *0.2* | *0.3* | |
| 2. Weighted co-occurrence counts (Unweighted * MCDIp). | | | | | |
| **Why** | 7 | 6 | 20 | 30 | 63 |
| **Where** | 70 | 60 | 2 | 3 | 135 |
| 3. Unweighted Sum/ Weighted Sum | | | | **Pro-KWo** | |
| **Why** | 63/220 | | = | 0.29 | |
| **Where** | 135/220 | | = | 0.61 | |
| All data analysis code found at: https://github.com/AzFlores/Pro-KWo | | | | | |

**Table.1** Hypothetical Pro-KWo scores for the words *why* and *where*. Co-Occurrence values are calculated within a 7 word forward moving window.

With these measures of children's vocabulary knowledge and key distributional measures of child-directed speech, we can explore the relationships between these vocabulary measures and distributional statistics of child-directed speech and build regression models that use the distributional measures to predict language outcomes. All analyses were performed in R. Mixed-effects logistic regression (glmer) analyses were performed with the lme4 package (version 1.1.26) (Bates, Maechler, Bolker, & Walker, 2015). Data and code are available at (https://github.com/AzFlores/Pro-KWo).

In our first set of analyses, we aimed to better understand the relationship between our four key distributional statistics (Frequency, Lexical Diversity, Document Diversity, Pro-KWo) and child language outcomes. We first computed correlations between our four distributional measures with each other, at each age. We then computed correlations between our four distributional measures and MCDIp at each age. Next, we build mixed-effects logistic regression models predicting our binary word production measure (1 = produced, 0 = did not produce) from our four distributional measures. We computed separate models at each age. We computed models with each distributional measure in isolation as well as together with the three other distributional measures to understand the predictive strength of each measure in isolation, as well as in the context of other predictors. All models contained random intercepts by child and by word.

Our second set of analyses mirror our first set of analyses, but we aimed to understand the role of grammatical class in moderating the relationship between our four distributional predictors and language outcomes. We first computed correlations between our four distributional measures separately for each of four grammatical classes (adjective, function word, noun, verb). We then computed correlations for each of our four distributional predictors with MCDIp, again, separately for each grammatical class, at each age. Finally, we build a mixed mixed-effects logistic regression model predicting our binary word production measure with our Pro-KWo measure. Crucially, we did not compute separate models for each grammatical class. Our goal was to better understand the prediction error across words of different grammatical classes in these regression models and illustrate that Pro-KWo is a measure robust to grammatical class, such that the regression model does not systematically exhibit greater prediction error for some grammatical classes.

**Results**

In order to better understand the relationship between each statistical predictor, we first examine the correlation between each predictor at 24 months of age (Figure 1). Within this age group, there exists a strong correlation between word frequency and both measures of document and lexical diversity. In addition, document and lexical diversity are themselves highly correlated. In contrast, our measure of Pro-KWo shows no significant correlation with either diversity measures and only a small correlation with word frequency. In order to better understand whether this pattern of relationships among our measures is consistent across development, we examined the correlations between our predictor variables in four additional age groups (Table 2). Across age groups we find that the magnitude of correlation coefficients among Pro-KWo and each of the other predictors is small. Despite being calculated with the same language corpus (child-available speech from the CHILDES corpus), Pro-KWo is generally not correlated with and is therefore likely accounting for different sources of variability than the other predictors investigated here. Incorporating information about the likelihood of a word being previously known to a measure derived from corpus statistics leads Pro-KWo to capture qualitatively different variance than the other distributional measures investigated here.

Next, we were interested in whether the four predictors showed a relationship to aggregate measures of children's vocabulary knowledge. We correlated each of the four predictors with the average proportion of children who produce that word (MCDIp) at each age group (Figure 2). Both frequency and lexical diversity show a small correlation with MCDIp, with the magnitude of correlation remaining largely consistent as age increases. In contrast, we found that Pro-KWo was moderately correlated with MCDIp at all age groups. Compared to other statistical predictors Pro-KWo showed the strongest relationship to MCDIp, with its effect increasing across age groups. We did not find any significant relationship between MCDIp and document diversity across any age group.

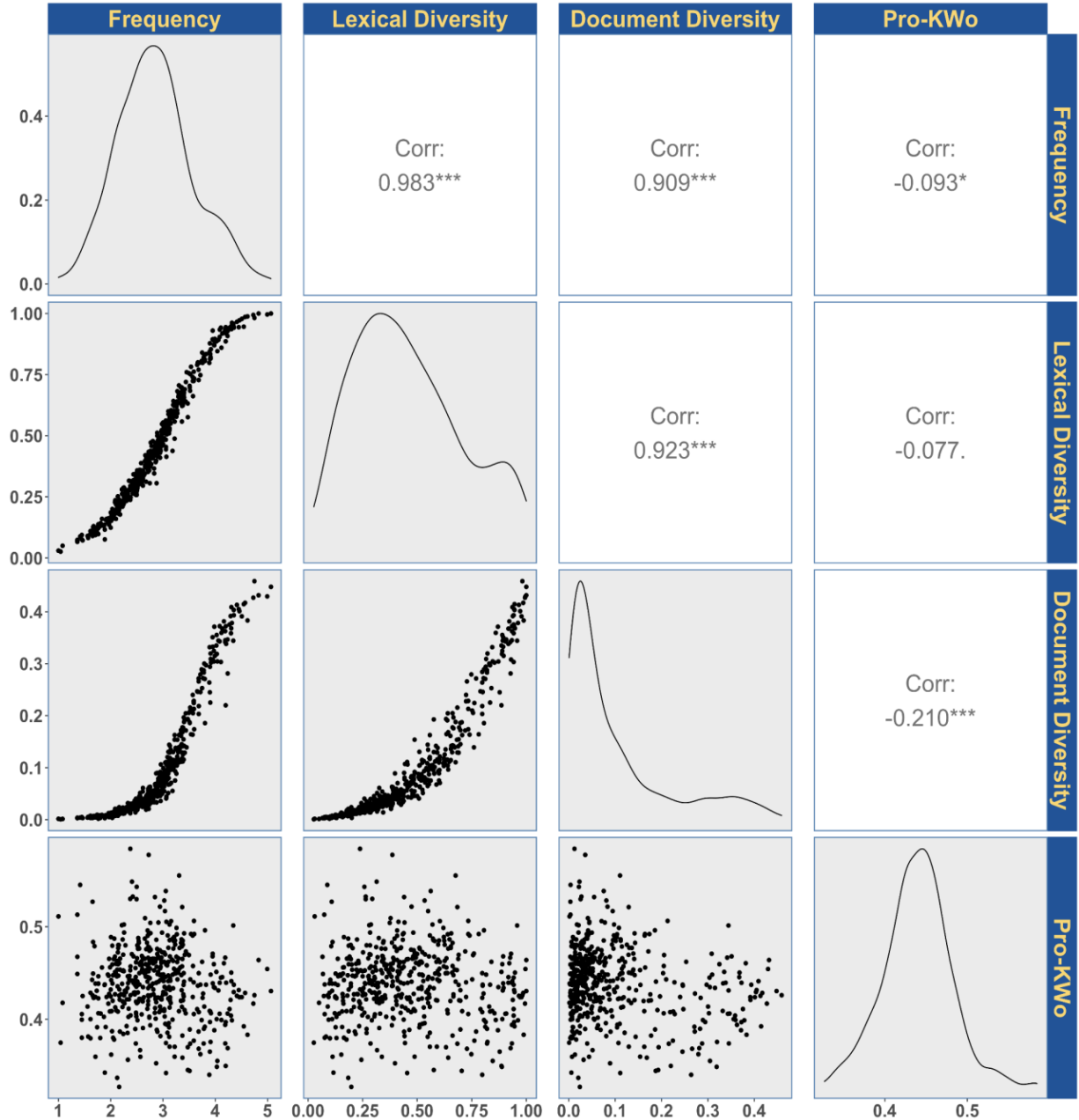

**Figure 1.** Correlogram of all distributional statistics at 24 months. *Frequency is the* $\log_{10}$ transformed cumulative frequency.

| | Frequency | | | | | Lexical Diversity | | | | | Document Diversity | | | | |
|---|---|---|---|---|---|---|---|---|---|---|---|---|---|---|---|
| | 18 | 21 | 24 | 27 | 30 | 18 | 21 | 24 | 27 | 30 | 18 | 21 | 24 | 27 | 30 |
| Frequency | 1.00 | 1.00 | 1.00 | 1.00 | 1.00 | | | | | | | | | | |
| Lexical Diversity | **0.97** | **0.97** | **0.98** | **0.98** | **0.98** | 1.00 | 1.00 | 1.00 | 1.00 | 1.00 | | | | | |
| Document Diversity | **0.89** | **0.89** | **0.90** | **0.91** | **0.91** | **0.94** | **0.94** | **0.92** | **0.91** | **0.89** | 1.00 | 1.00 | 1.00 | 1.00 | 1.00 |
| Pro-KWo | **-0.13** | -0.07 | **-0.09** | -0.05 | -0.06 | **-0.12** | -0.07 | -0.07 | -0.04 | -0.05 | **-0.22** | **-0.19** | **-0.21** | **-0.17** | **-0.17** |

All data analysis code found at: https://github.com/AzFlores/Pro-KWo

**Table 2.** Correlation of all distributional statistics across five age groups, but computed within each age. For example, the correlation of frequency at 18 months with Pro-KWo at 18 months is 0.13, at 21 months is -0.07, at 24 month is -0.09, etc. Bolded signifies the correlation is significant at the 0.01 level (2-tailed).

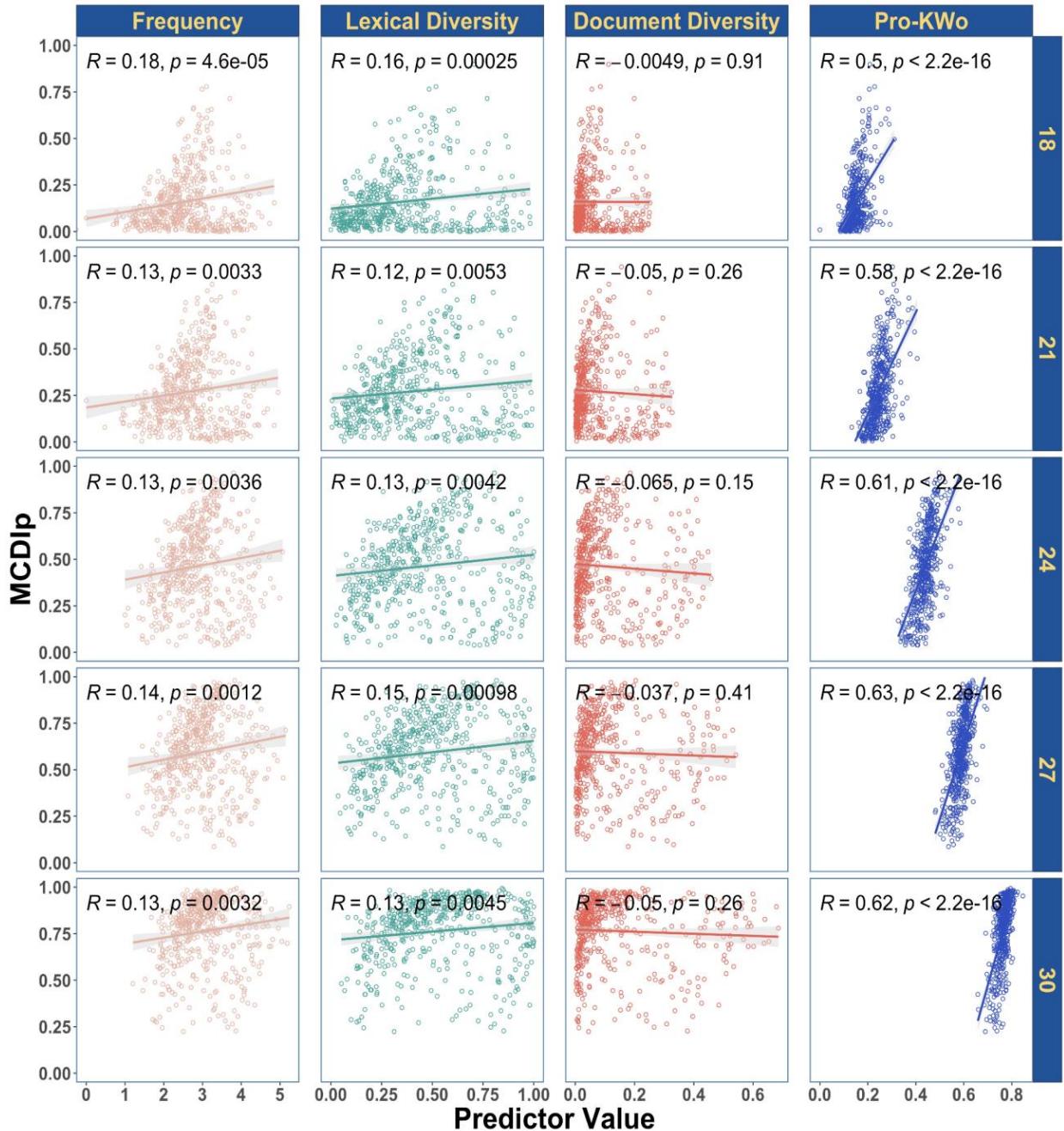

**Figure 2.** Correlation between each distributional statistic and MCDIp across age groups. Frequency is the $\log_{10}$ transformed cumulative frequency. Pro-KWo Shuffle correlations for ages 18(-0.04), 21(-0.07), 24(-0.04), 27(-0.01), 30(-0.03) were non-significant.

Up to this point we have examined how aggregate measures of children's vocabulary knowledge (MCDIp) relate to each of our predictor variables. These analyses allow us to see normative trends in the data, and understand the relationships between our predictors, but do not allow us to capture systematic patterns within individuals in the sets of words that are known or not known. To capture individual variability in patterns of known words, we must predict whether an individual child does or does not know a word at a given age. Doing so would allow us to draw conclusions regarding how each of our predictors can be used to capture which words are more or less likely to be produced by children at different ages.

In order to investigate how each of the four predictors is able to predict binary production data, we fit a separate multilevel logistic regression model using each of our four distributional statistics as predictors (standardized and centered), at each age with each child's individual binary production data as the outcome variable. We included random intercepts of both individual participants and individual words to account for the inherently structured nature of our dataset. In Figure 3 we show individual fixed effect estimates, where each point represents a separate multilevel model.

Our results suggest that there are developmentally sensitive differences in how predictive each measure is, for instance document diversity was found to be a significant predictor within the early age groups, but not in the latter age groups. The opposite was true for lexical diversity where the proportion of MCDI word types that each word co-occurred with was a significant predictor at later age groups but not early. Across all age groups, Pro-KWo was a strong predictor of the binary production data, and its effect increases with age. Finally, frequency showed a significant effect across all age groups except at 21 months. A full account of the modeling results for the single predictor models can be found in Table 3.

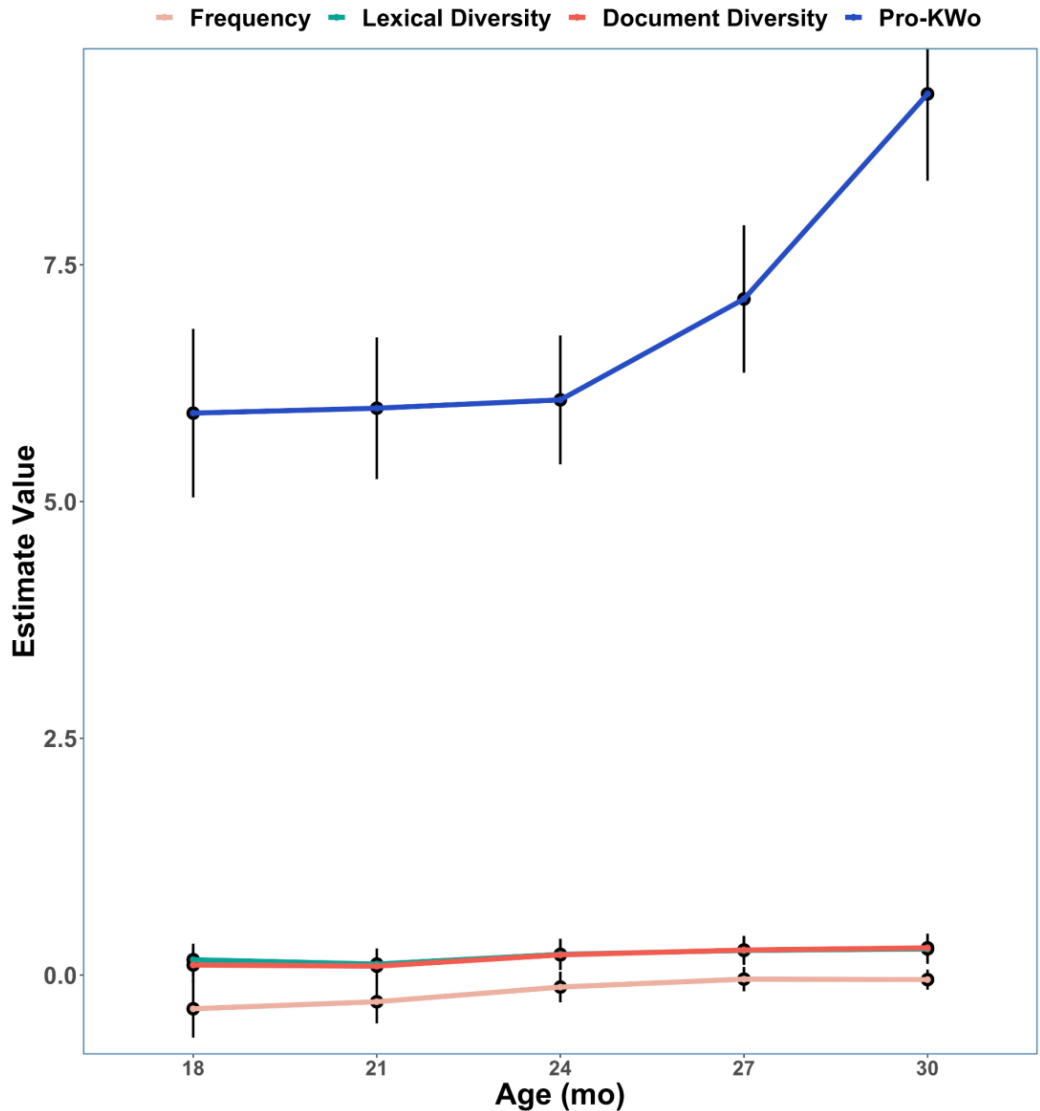

**Figure 3.** Fixed effect estimates for single predictor models. Each point represents a
single model with 95% confidence intervals around each estimate.

In addition to comparing single predictor models to each other, we also fit multilevel models that included all of the predictors at each age (Table 4). This allows us to assess how predictive each measure is in the presence of the other three predictors.

When comparing the fixed effects across our single predictor models (Table 3) to a model with all predictors (Table 4), we find that the effects of frequency, document and lexical diversity are all smaller in the single predictor models. Adding additional predictors to the model increases in variance accounted for by frequency, document and lexical diversity, suggesting that these predictors are accounting for different, unshared sources of variance. In contrast, unlike the other three predictors, we find that fixed effect estimates for Pro-KWo remain high and *decrease* only slightly in the multiple predictor models relative to the single predictor models. Given the consistency of Pro-KWo as a predictor across both the single predictor and full models, this measure may be capturing both variability that is captured by the other variables, *plus* other unique sources of variability that the other predictors fail to capture.

**Mixed Effects Models with Single Predictors**

Production ~ *Predictor* + (1|Subject) + (1|Word)

|  | Estimate | Std. Error | z-value | Pr(>|z|) | 2.50% | 97.50% |
|---|---|---|---|---|---|---|
| **Frequency** | | | | | | |
| 18 | 0.162 | 0.086 | 1.877 | 0.061 | -0.007 | 0.331 |
| 21 | 0.116 | 0.084 | 1.380 | 0.168 | -0.049 | 0.282 |
| 24 | 0.219 | 0.084 | 2.605 | 0.009 | 0.054 | 0.384 |
| 27 | 0.260 | 0.078 | 3.342 | 0.001 | 0.108 | 0.413 |
| 30 | 0.276 | 0.081 | 3.427 | 0.001 | 0.118 | 0.435 |
| **Lexical Diversity** | | | | | | |
| 18 | 0.106 | 0.094 | 1.125 | 0.261 | -0.079 | 0.291 |
| 21 | 0.094 | 0.086 | 1.092 | 0.275 | -0.075 | 0.263 |
| 24 | 0.213 | 0.082 | 2.605 | 0.009 | 0.053 | 0.373 |
| 27 | 0.265 | 0.074 | 3.558 | 0.000 | 0.119 | 0.410 |
| 30 | 0.287 | 0.077 | 3.737 | 0.000 | 0.136 | 0.437 |
| **Document Diversity** | | | | | | |
| 18 | -0.356 | 0.156 | -2.282 | 0.023 | -0.661 | -0.050 |
| 21 | -0.282 | 0.116 | -2.426 | 0.015 | -0.511 | -0.054 |
| 24 | -0.127 | 0.083 | -1.526 | 0.127 | -0.289 | 0.036 |
| 27 | -0.043 | 0.066 | -0.065 | 0.517 | -0.172 | 0.086 |
| 30 | -0.048 | 0.054 | -0.881 | 0.378 | -0.154 | 0.058 |
| **Pro-KWo** | | | | | | |
| 18 | 5.934 | 0.454 | 13.062 | 0.000 | 5.044 | 6.825 |
| 21 | 5.986 | 0.382 | 15.670 | 0.000 | 5.237 | 6.734 |
| 24 | 6.074 | 0.347 | 17.484 | 0.000 | 5.393 | 6.755 |
| 27 | 7.139 | 0.397 | 17.991 | 0.000 | 6.361 | 7.917 |
| 30 | 9.305 | 0.468 | 19.869 | 0.000 | 8.387 | 10.223 |

All data analysis code found at: https://github.com/AzFlores/Pro-KWo

**Table 3.** Parameter estimates for single predictor models. For all models, random intercepts of participants and words are included.

**Mixed Effects Models with All Predictors**

Production ~ *Frequency + Lexical Diversity + Document Diversity + Pro-KWo* + (1|Subject) + (1|Word)

|  | Estimate | Std. Error | z-value | Pr(>|z|) | 2.50% | 97.50% |
|---|---|---|---|---|---|---|
| **18** | | | | | | |
| Frequency | 0.300 | 0.301 | 0.998 | 0.318 | -0.289 | 0.889 |
| Lexical Diversity | 1.945 | 0.474 | 4.105 | 0.000 | 1.016 | 2.873 |
| Document Diversity | -3.548 | 0.427 | -8.310 | 0.000 | -4.385 | -2.711 |
| Pro-KWo | 4.902 | 0.447 | 10.955 | 0.000 | 4.025 | 5.779 |
| **21** | | | | | | |
| Frequency | 0.193 | 0.306 | 0.631 | 0.528 | -0.407 | 0.794 |
| Lexical Diversity | 1.424 | 0.413 | 3.449 | 0.001 | 0.615 | 2.233 |
| Document Diversity | -2.079 | 0.288 | -7.220 | 0.000 | -2.644 | -1.515 |
| Pro-KWo | 5.078 | 0.382 | 13.303 | 0.000 | 4.330 | 5.826 |
| **24** | | | | | | |
| Frequency | 0.871 | 0.346 | 2.514 | 0.012 | 0.192 | 1.550 |
| Lexical Diversity | 0.628 | 0.367 | 1.710 | 0.087 | -0.092 | 1.349 |
| Document Diversity | -1.278 | 0.168 | -7.615 | 0.000 | -1.607 | -0.949 |
| Pro-KWo | 5.369 | 0.353 | 15.212 | 0.000 | 4.677 | 6.060 |
| **27** | | | | | | |
| Frequency | 0.824 | 0.326 | 2.529 | 0.011 | 0.185 | 1.462 |
| Lexical Diversity | 0.500 | 0.314 | 1.592 | 0.111 | -0.116 | 1.115 |
| Document Diversity | -0.934 | 0.128 | -7.296 | 0.000 | -1.184 | -0.683 |
| Pro-KWo | 6.318 | 0.396 | 15.965 | 0.000 | 5.542 | 7.093 |
| **30** | | | | | | |
| Frequency | 1.226 | 0.344 | 3.561 | 0.000 | 0.551 | 1.901 |
| Lexical Diversity | 0.366 | 0.295 | 1.238 | 0.216 | -0.213 | 0.945 |
| Document Diversity | -0.915 | 0.102 | -8.942 | 0.000 | -1.116 | -0.715 |
| Pro-KWo | 8.116 | 0.490 | 16.564 | 0.000 | 7.156 | 9.076 |

All data analysis code found at: https://github.com/AzFlores/Pro-KWo

**Table 4.** Parameter estimates for all predictor models. For all models, random intercepts of participants and words are included.

**Grammatical Class**

In many statistical models of word learning, the effect of distributional statistics on word knowledge have been shown to differ based on grammatical class, with smaller effect sizes found when examining across all words, and larger effect sizes when examining words within a specific grammatical class. Different patterns across grammatical classes might suggest that different learning mechanisms underlie the learning of different categories, so they cannot be merged together in statistical analyses-- in short, the same predictors do not operate similarly across grammatical category. We were interested in whether we would similarly find larger effects within than between grammatical classes for Pro-KWo, or if this measure would be robust to the grammatical category, suggesting a common learning mechanism.

We examine the same 24 month dataset from Figure 1, calculating the relationships between our four predictors, with the addition of each word's grammatical class (Figure 4). Measures of frequency, document and lexical diversity are highly correlated with each other both across all words and within each grammatical class. This is not the case when examining the relationship between Pro-KWo and the other distributional statistics. When aggregating across all grammatical classes, there appeared to be no relationship between Pro-KWo and both frequency and lexical diversity across all words (i.e pearson's r of -0.093 and -0.077 respectively), however within each grammatical class there are significant moderate correlations, particularly within nouns and function words. When examining the relationship between Pro-KWo and document diversity we find a stronger correlation across all words compared to within grammatical categories, with the exception of function words.

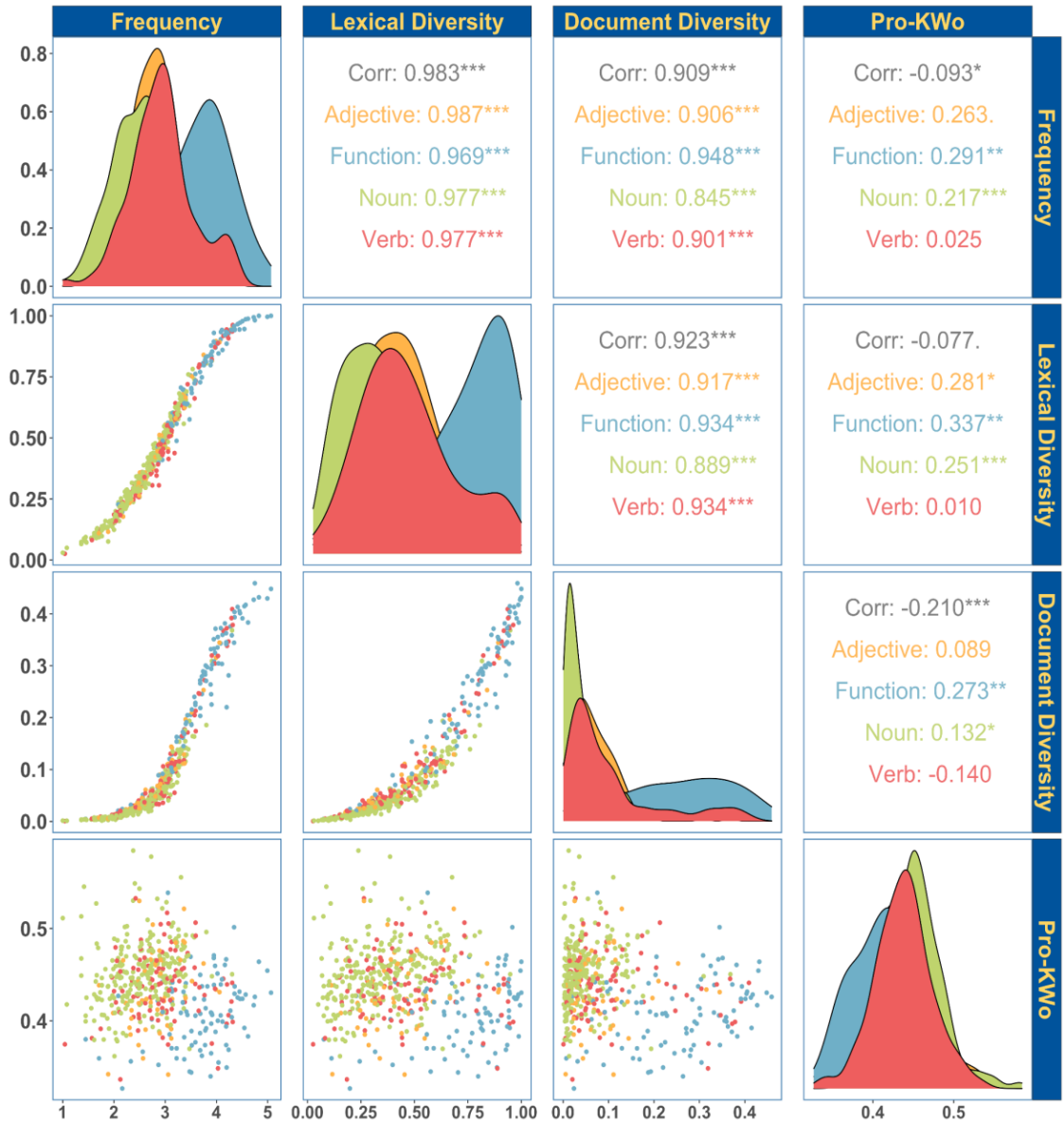

**Figure 4.** Correlogram of all distributional statistics at 24 months within grammatical class. *Frequency is the* $\log_{10}$ transformed cumulative frequency.

The low correlation across grammatical class suggests that whereas in aggregate Pro-KWo and the other three predictors seemed to capture different variance in aggregate, when broken down by category, Pro-KWo and other predictors seem to capture similar sources of variance. In essence, some feature of what makes a noun different from an adjective from a verb such that frequency, lexical and document diversity only account for variability *within* class, is already incorporated into the Pro-KWo measure. The sets of words, especially known words, that items in different grammatical classes co-occur with seems to systematically vary by grammatical class in important ways that leads Pro-KWo to be a robust predictor across grammatical class. These findings suggest that there is still more to be understood regarding how prior knowledge interacts with the learning of words of specific grammatical classes. Further it suggests that Pro-KWo as a measure of the quality of linguistic contexts that children hear may not be fully independent of other predictors as there are small relationships between Pro-KWo and the other four distributional predictors when taking into account grammatical class.

Next, we examined each of our statistical predictor's relationship to MCDIp scores, both across and within grammatical category. Figure 5 illustrates the relationship between each statistical predictor and MCDIp at 24 months, with different colors depicting different grammatical categories. Figure 6 illustrates the pearson correlation coefficient of each of our statistical predictor's relationship to MCDIp scores across age groups. Again, different colors depicting different grammatical categories. In Figure 5, words of the same grammatical class tend to cluster together, for frequency, lexical diversity and document diversity, but not Pro-KWo, where scores are more homogeneously distributed across a word's grammatical class. The difference in correlation by grammatical class is particularly evident in Figure 6, where the magnitude of the correlation varies by grammatical class (and is somewhat consistent across age). However, while for frequency, lexical diversity and document diversity, correlations are substantially lower when all grammatical classes are aggregated, for Pro-KWo the correlation remains high when aggregating across grammatical class.

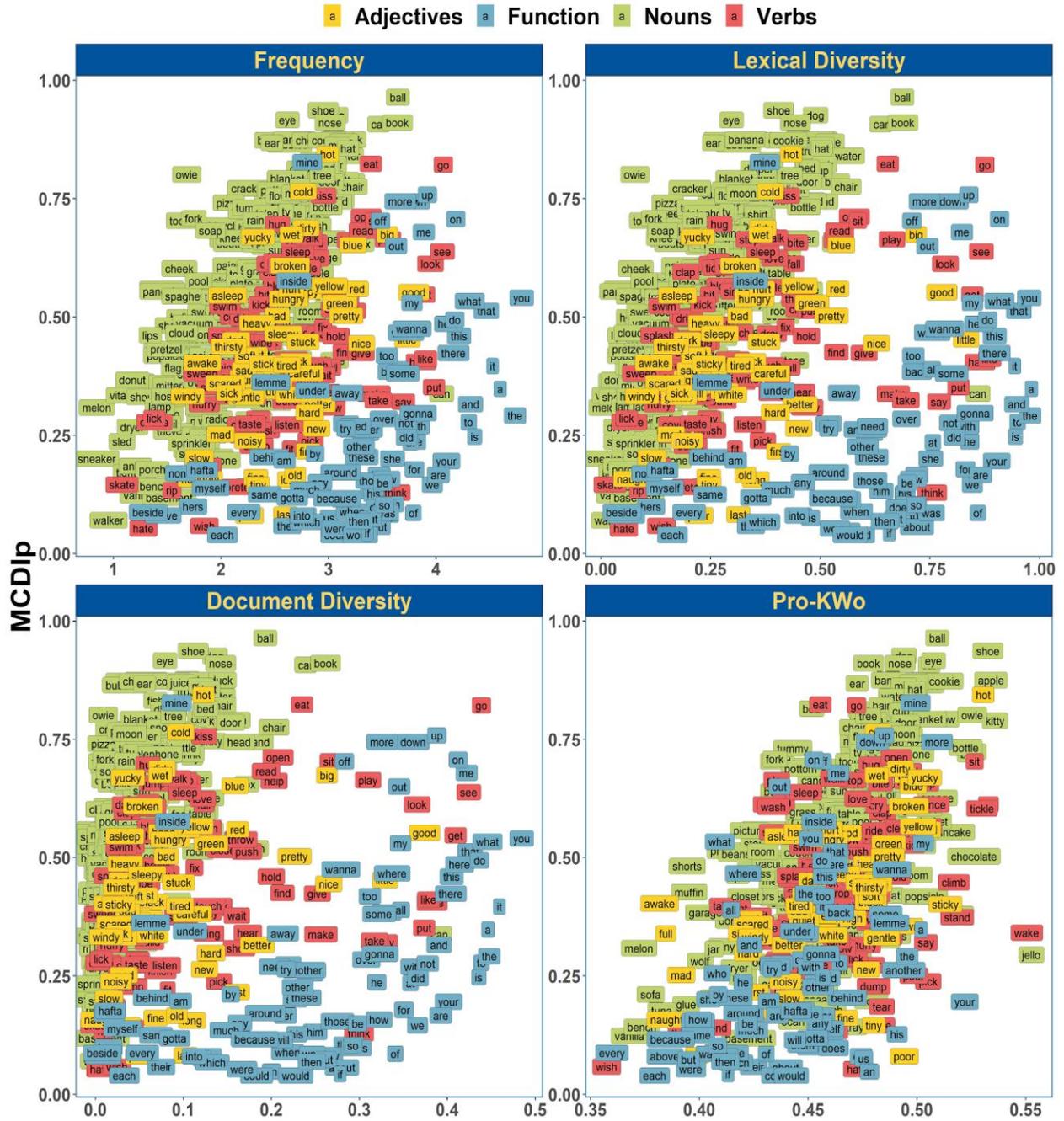

**Figure 5.** Correlation of MCDIp and Distributional Statistics at 24 months.

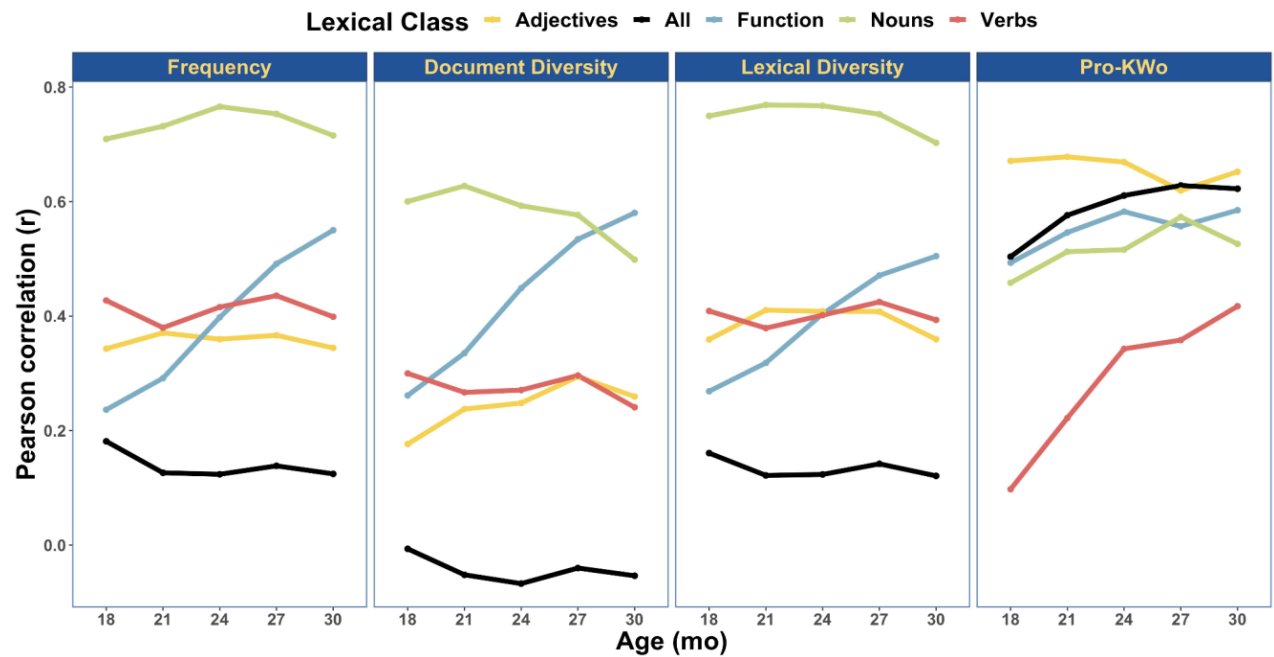

**Figure 6.** Correlation of MCDIp and distributional statistics across age for each word's grammatical class.

Across age groups, a consistent pattern emerges such that measures of frequency, document and lexical diversity show only a small correlation with MCDIp across all words. However, when taking into account grammatical categories we see a marked increase in correlation values. This increase is once again most noticeable for nouns. In contrast, Pro-KWo shows a qualitatively different pattern than the other predictors and maintains an overall consistent relationship with MCDIp. The only exception to the consistent relationship between Pro-KWo and MCDIp within and cross grammatical class is that Pro-KWo shows a smaller correlation when examining only verbs. This consistent pattern across grammatical class is an important difference between Pro-KWo and other statistical predictors of word learning. Most other models of word learning find stronger effects within grammatical categories than across all grammatical categories but Pro-KWo shows no similar boost: it is as strong a predictor within a single grammatical category or across all words.

The lack of a grammatical class effect here is important. Previous assessments have typically needed to take into account a word's grammatical class in order to predict learning outcomes, implying that various predictors such as frequency, lexical and document diversity exert different effects on words of different grammatical classes. The separation of the dataset by grammatical class is not necessary for Pro-KWo. Whereas other statistical models of word learning needed to attribute independent variance to grammatical class, with Pro-KWo no such attribution is necessary. The same predictor accounts for variability in word learning similarly across grammatical class suggesting that variance that was previously attributed to grammatical class may indeed stem from systematic differences in the distributional patterns across grammatical class, not grammatical class per-se. This finding is consistent with all of the behavioral work showing that general learning mechanisms can help children learn the meanings of words across a wide range of grammatical classes (Arias-Trejo & Alva, 2013; Christophe et al., 2013; Fisher, Gertner, Scott, Yuan, 2010; Gentner, 1989; Gentner & Namy, 2006; Hills et al., 2010; Huebner & Willits, 2018; Lany & Saffran, 2010, 2013; Naigles, 1990; Sadeghi, Scheutz, Krause, 2017; Wojcik and Saffran, 2013).

**General Discussion**

In the current work, we aimed to incorporate insights from behavioral findings into statistical models of language learning. Specifically, we aimed to incorporate the consistency of learning mechanisms across grammatical class and the role of linguistic quality of word learning contexts-- specifically the contexts that allow learners to leverage their existing world knowledge to learn new words. We introduced a new metric (Pro-KWo) which is a co-occurrence based statistical predictor that integrates information about the likelihood with which children know those words in the statistical distribution. This measure accounts for more variability in word learning than other distributional measures, but crucially accounts for variability both within and across grammatical class.

We evaluated the extent to which four distributional measures, three well-investigated measures: word frequency, lexical diversity and document diversity and our new Pro-KWo measure, predicted vocabulary outcomes within and across a word's grammatical class. This work makes two important contributions. First, consistent with existing findings, our results showed that measures of a word's frequency, lexical diversity and document diversity are all highly correlated with aggregate measures of children's vocabulary knowledge within but not across grammatical class. However, Pro-KWo was a robust predictor of word knowledge not only within each grammatical class but also across all words aggregated together. Second, by computing measures of word knowledge and distributional predictors separately for each age group, we were able to investigate how the distributional measures predicted word knowledge differently over development. Rather than being uniformly predictive over time, different distributional measures accounted for more or less variability at different child ages.

Unlike other predictors of word knowledge, Pro-KWo was robust across all grammatical classes. Rather than posit different effects of Pro-KWo by grammatical class, we found that this measure was uniformly predictive, suggesting that the underlying mechanism or mechanisms that are indexed by Pro-KWo may also act uniformly across grammatical class, removing the need to posit a different learning mechanism or mechanisms by grammatical class. To better understand the nature of the predictiveness of Pro-KWo across grammatical class, we dug deeper into the prediction error across items of Pro-KWo. We fit logistic regression models predicting binary (produced/not produced) vocabulary outcomes with Pro-KWo as the single predictor (Figure 7) at 24 months. We then computed the prediction error made by the model for each item (word). Then we correlated the model's prediction error of each word with the aggregate measure of vocabulary knowledge to yield the relationship between the degree of prediction error relative to the proportion of children who produced a particular item. Figure 7 shows that the prediction error of each grammatical class is largely overlapping, suggesting that Pro-KWo is not systematically under- or over-estimating production likelihood of grammatical class. However, it is also clear from Figure 7 that there is some clustering by grammatical class. Some nouns are clustered above the regression line at the top while some function words are clustered at the bottom. So Pro-KWo, while relatively more robust to grammatical class than other predictions, still shows some evidence of small effects of grammatical class. Pro-KWo is *under*-estimating the likelihood that children produce some nouns and *over*-estimating the likelihood that children produce some function words. Of course, it is not clear if this over and underestimation is an effect of grammatical class per se, or Pro-KWo is broadly underestimating the most frequently produced words and overestimating the least frequently produced words, which happen to be nouns and function words respectively. Or perhaps the effect of Pro-KWo should be modeled non-linearly to best account for extreme values. Nevertheless, we find that Pro-KWo retains some degree of sensitivity to grammatical class, though far less than other distributional predictions.

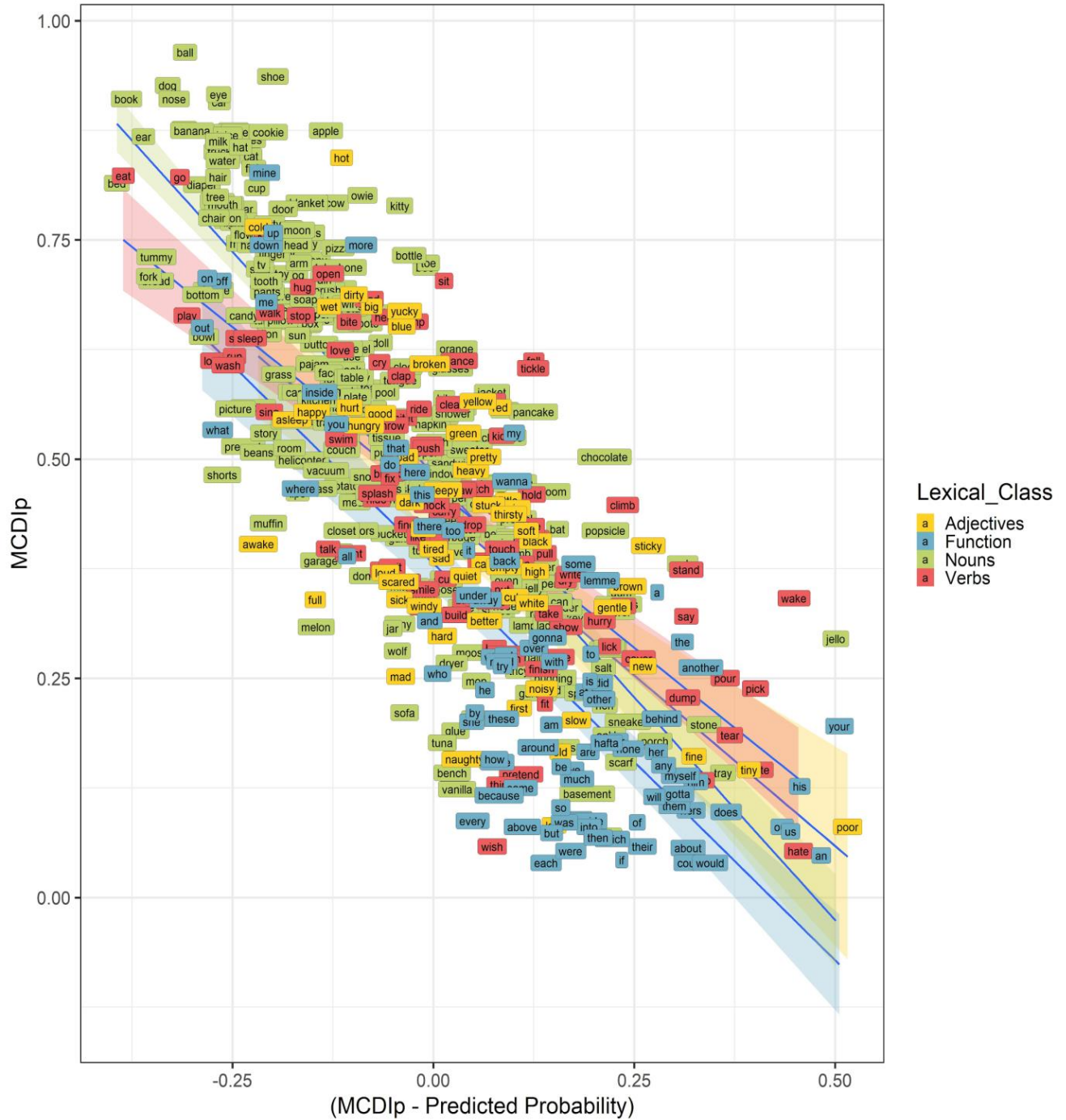

**Figure 7.** Correlation of MCDIp with predicted model probabilities at 24 months. Positive values along the x-axis represent over predictions, while negative values represent under predicted values.

*Model (logistic regression):* Production ~ Pro-KWo

An important question is what exactly Pro-KWo might be capturing that allows it to robustly predict word knowledge across grammatical class. Pro-KWo is clearly incorporating some feature that in other statistical models is captured by separately modeling grammatical class. Our hypothesis is that words in different grammatical classes are systematically more or less likely to occur alongside words known to children. So it is systematic distributional differences across grammatical class, not grammatical class per se, that leads other statistical predictors of word learning to fail to account for learning across grammatical class. Because Pro-KWo captures these systematic distributional differences, it is robust across grammatical class.

Another insight from this work comes from leveraging the existence of cross-sectional corpus data and vocabulary inventories to compute the effects of distributional predictors. First, this approach allows us to conceive of distributional predictors not as a fixed value, but values that change as children's cumulative language input changes over time. Caregiver input to children changes with age (Fenson, 2007, Huebner & Willits, 2021), so this cross-sectional approach allows us to capture how distributional properties of child-directed utterances change over time, and compute distributional statistics that reflect these changing statistics over time. Second, the cross-sectional approach allows us to see the relative variability accounted for by different statistical predictors over time. For example, document diversity accounted for the most variance in word knowledge at younger ages, and Pro-KWo accounted for more variability at older ages. It is not clear if these changes over time reflect changes in distributional properties of child-directed and available speech over time, differences in language learning processes as children have more existing language knowledge to aid subsequent learning, differences in which words and phrases children are able to or motivated to produce, or another factor. We believe this cross-sectional approach suggests various lines of future work and unanswered questions into which both statistical models and behavioral approaches may be able to provide insight.

Overall our results provide evidence that previously proposed learning mechanisms and biases which have historically focused on nouns, may extend to words of other grammatical classes. To our knowledge, Pro-KWo is the first statistical predictor of word learning that does not interact with a word's grammatical class. Pro-KWo then represents a first step in characterizing what kinds of information can be used to quantify which language experiences may be most useful for learning new words. And while the current implementation defines prior knowledge as an aggregate measure of children's productive vocabulary, it is a measure that closely approximates the quality of speech children hear in a way not previously reported. There is reason to believe more refined and targeted accounts of children's prior knowledge may be even more useful when incorporated into a distributional predictor of word learning. As has been noted in prior behavioral work on word learning there are additional factors beyond prior knowledge which account for vocabulary outcomes. And achieving a way of capturing these factors and subsequently incorporating them into a distributional statistic may provide more ways in which distributional statistics can be used to study word learning. For instance, a great deal of research has identified that language episodes in which the child and parent are both jointly attending to a referent are particularly informative and promote word learning. Finding ways of identifying episodes of joint attention from speech corpora may not be a straightforward process. However, such efforts may be worthwhile in that they begin to further increase the utility of large naturalistic datasets by adding important meta-information by which to weight language statistics. Further we contend that such approaches are necessary in order to increase the overall validity of distributional statistics of language learning.